\documentclass{article}
\usepackage{spconf,amsmath,graphicx}
\usepackage{multirow}
\usepackage{multicol}
\usepackage{booktabs}
\usepackage{xcolor}
\usepackage{float}

\graphicspath{
{./figures/}
}

\title{Semantic Sentence Embeddings for Paraphrasing and Text Summarization}
\name{Chi Zhang, 
      Shagan Sah,
      Thang Nguyen,
      Dheeraj Peri,
      Alexander Loui$^{\dagger}$,
      Carl Salvaggio,
      Raymond Ptucha
}
\address{ Rochester Institute of Technology, Rochester, NY 14623, USA \\
    $^{\dagger}$ Kodak Alaris Imaging Science R\&D, Rochester, NY 14615, USA} 
\begin{document}
%
\maketitle
\begin{abstract}
This paper introduces a sentence to vector encoding framework suitable for advanced natural language processing.  Our latent representation is shown to encode sentences with common semantic information with similar vector representations. The vector representation is extracted from an encoder-decoder model which is trained on sentence paraphrase pairs.  We demonstrate the application of the sentence representations for two different tasks -- sentence paraphrasing and paragraph summarization, making it attractive for commonly used recurrent frameworks that process text. Experimental results help gain insight how vector representations are suitable for advanced language embedding.
\end{abstract}
\begin{keywords}
sentence embedding, sentence encoding, sentence paraphrasing, text summarization, deep learning
\end{keywords}
%
\section{Introduction}
\label{sec: intro}

Modeling temporal sequences of patterns requires the embedding of each pattern into a vector space.  For example, by passing each frame of a video through a Convolutional Neural Network (CNN), a sequence of vectors can be obtained.  These vectors are fed into a Recurrent Neural Network (RNN) to form a powerful descriptor for video annotation \cite{venugopalan2015translating,yao2015describing,venugopalan15iccv}. Similar, techniques such as word2vec \cite{mikolov2013distributed} and GloVe \cite{pennington2014glove} have been used to form vector representations of words. Using such embeddings, sentences become a sequence of word vectors. When these vector sequences are fed into a RNN, we get a powerful descriptor of a sentence \cite{cho2014properties}.

Given vector representations of a sentence and video, the mapping between these vector spaces can be solved, forming a connection between visual and textual spaces.  This enables tasks such as captioning, summarizing, and searching of images and video to become more intuitive for humans. By vectorizing paragraphs \cite{le2014distributed}, similar methods can be used for richer textual descriptions.

Recent advances at vectorizing sentences represent exact sentences faithfully \cite{kalchbrenner2014convolutional,le2014distributed,zhao2015self}, or pair a current sentence with prior and next sentence \cite{kiros2015skip}.  Just like word2vec and GloVe map words of similar meaning close to one another, we desire a method to map sentences of similar meaning close to one another.  For example, the toy sentences ``A man jumped over the stream'' and ``A person hurdled the creek'' have similar meaning to humans, but are not close in traditional sentence vector representations.  Just like the words flower, rose, and tulip are close in good sentence to vector representations, our toy sentences must lie close in the introduced embedded vector space.

Inspired by the METEOR \cite{banerjee2005meteor} captioning benchmark which allows substitution of similar words, we choose to map similar sentences as close as possible. We utilize both paraphrase datasets and ground truth captions from multi-human captioning datasets.  For example, the MS-COCO dataset \cite{lin2014microsoft} has over 120K images, each with five captions from five different evaluators. On average, each of these five captions from each image should convey the same semantic meaning.  We present an encoder-decoder framework for sentence paraphrases, and generate the vector representation of sentences from this framework which maps sentences of similar semantic meaning nearby in the vector encoding space.

The main contributions of this paper are 1) The usage of sentences from widely available image and video captioning datasets to form sentence paraphrase pairs, whereby these pairs are used to train the encoder-decoder model; 2) We show the application of the sentence embeddings for paragraph summarization and sentence paraphrasing, whereby evaluations are performed using metrics, vector visualizations and qualitative human evaluation; and 3) We extend the vectorized sentence approach to a hierarchical architecture, enabling the encoding of more complex structures such as paragraphs for applications such as text summarization.

The rest of this paper is organized as follows: Section \ref{sec:related} reviews some related techniques. Section \ref{sec:method} presents the proposed encoder-decoder framework for sentence and paragraph paraphrasing. Section \ref{sec:result} discusses the experimental results. Concluding remarks are presented in Section \ref{sec:conclusion}.

\section{Related Work}
\label{sec:related}

Most machine learning algorithms require inputs to be represented by fixed-length feature vectors. This is a challenging task when the inputs are text sentences and paragraphs. Many studies have addressed this problem in both supervised and unsupervised approaches. For example, \cite{kiros2015skip} presented a sentence vector representation while \cite{le2014distributed} created a paragraph vector representation. An application of such representations is shown by \cite{choi2017textually}, that has used individual sentence embeddings from a paragraph to search for relevant video segments.

An alternate approach uses an encoder-decoder \cite{sutskever2014sequence} framework that first encodes ${f}$ inputs, one at a time to the first layer of a two layer Long Short-Term Memory (LSTM), where ${f}$ can be of variable length.  Such an approach is shown for video captioning tasks by S2VT \cite{venugopalan2015sequence} that encodes the entire video, then decodes one word at a time.

There are numerous recent works on generating long textual paragraph summaries from videos. For example, \cite{yu2016video} present a hierarchical recurrent network that comprise of a paragraph generator that is built on top of a sentence. The sentence generator encodes sentences into compact representations and the paragraph generator captures inter-sentence dependencies. \cite{shin2016beyond} performed similar narratives from long videos by combining sentences using connective words at appropriate transitions learned using unsupervised learning.

\section{Methodology}
\label{sec:method}

The vector representation of a sentence is extracted from an encoder-decoder model on sentence paraphrasing, and then tested on a text summarizer.

\subsection{Vector Representation of Sentences} 
\label{subsec:paraphrase}

\begin{figure}[!ht]
  \centering
  \centerline{\includegraphics[width=.45\textwidth,trim=2 2 0 0,clip]{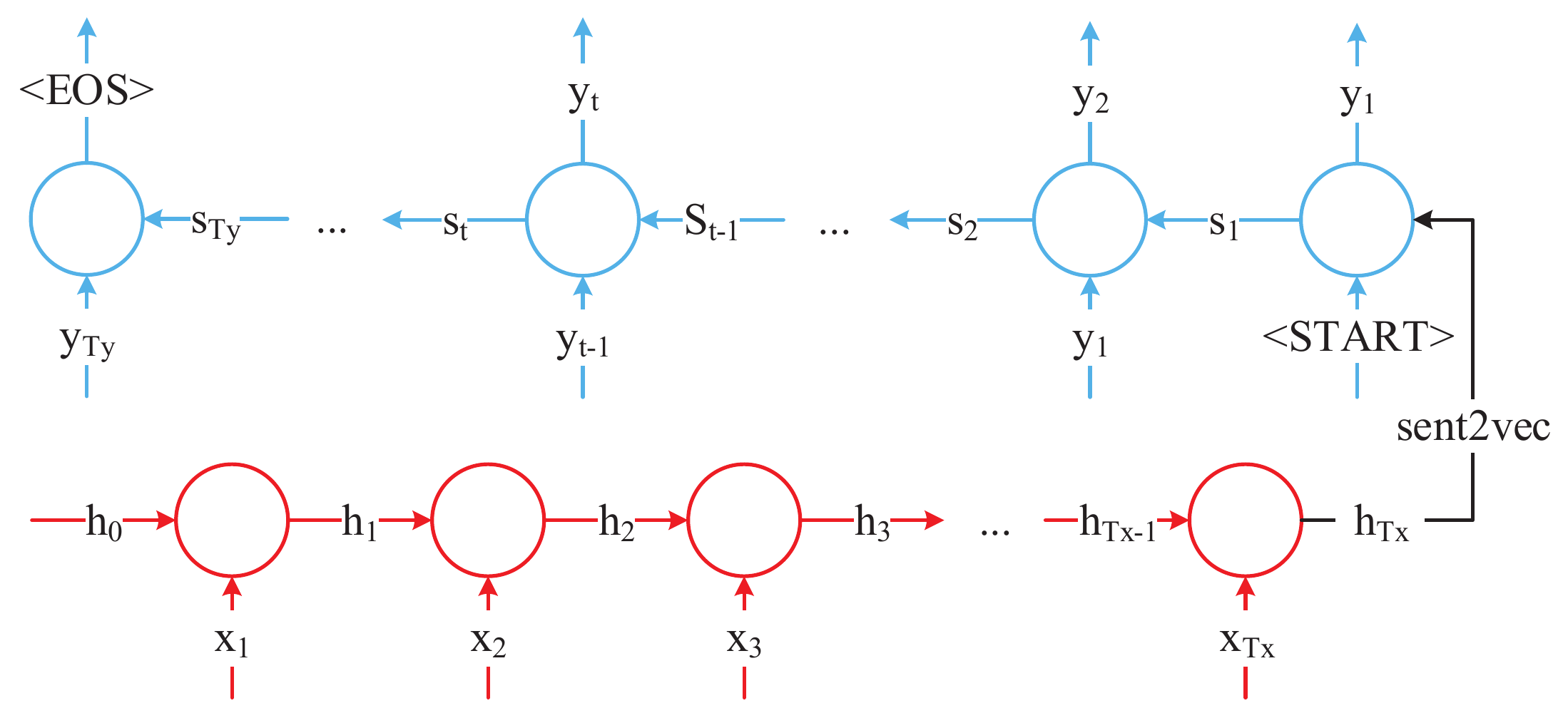}}
  \vspace{-.1in}
  \caption{The sentence paraphrasing model. The red and blue cells represent encoder and decoder respectively. The intermediate vector in black (sent2vec) is vector encoded sentence.}
\label{fig:paraphrase}
\end{figure}

We consider the sentence paraphrasing framework as an encoder-decoder model, as shown in Fig. \ref{fig:paraphrase}. Given a sentence, the encoder maps the sentence into a vector (sent2vec) and this vector is fed into the decoder to produce a paraphrase sentence. We represent the paraphrase sentence pairs as $(S_x, S_y)$. Let $x_i$ denote the word embedding for sentence $S_x$; and $y_j$ denote the word embedding for sentence $S_y$, $i \in \{1...T_x\}, j \in \{1...T_y\}$ where $T_x$ and $T_y$ are the length of the paraphrase sentences.

Several choices for encoder have been explored, including LSTM, GRU \cite{chung2014empirical} and BN-RHN \cite{zhang2017batch}. In our model, we use an RNN encoder with LSTM cells since it is easy to be implemented and performs well on this model. Specifically, the words in $S_x$ are converted into token IDs and then embedded using GloVe \cite{pennington2014glove}. To encode a sentence, the embedded words are iteratively processed by the LSTM cell \cite{sutskever2014sequence}.

The decoder is a neural language model which conditions on the encoder output $h_{T_x}$. The computation is similar to that of the encoder. The vector $h_{T_x}$ encodes the input sentence into a vector and is known as the vector representation of the input sentence, or ``sent2vec'' in this paper. Note that we do not have attention between encoder and decoder. This ensures that all the information extracted from the input sentence by encoder goes through sent2vec. In other words, attention is not adopted in order to avoid information leakage.

In full softmax training, for every training example in the word level, we would need to compute logits for all classes. However, this can get expensive if the universe of classes depending on size of vocabulary is very large. Given the predicted sentence and ground truth sentence, we use sampled softmax \cite{JeanCMB14} as a candidate sampling algorithm. For each training sample, we pick a small set of sampled classes according to a chosen sampling function. A set of candidates $C$ is created containing the union of the target class and the sampled classes. The training task figures out, given this set $C$, which of the classes in $C$ is the target class.

\subsection{Hierarchical Encoder for Text Summarization}
\label{subsec:summarize}

\begin{figure}[!ht]
  \centering
  \centerline{\includegraphics[width=.45\textwidth,trim=2 2 0 0,clip]{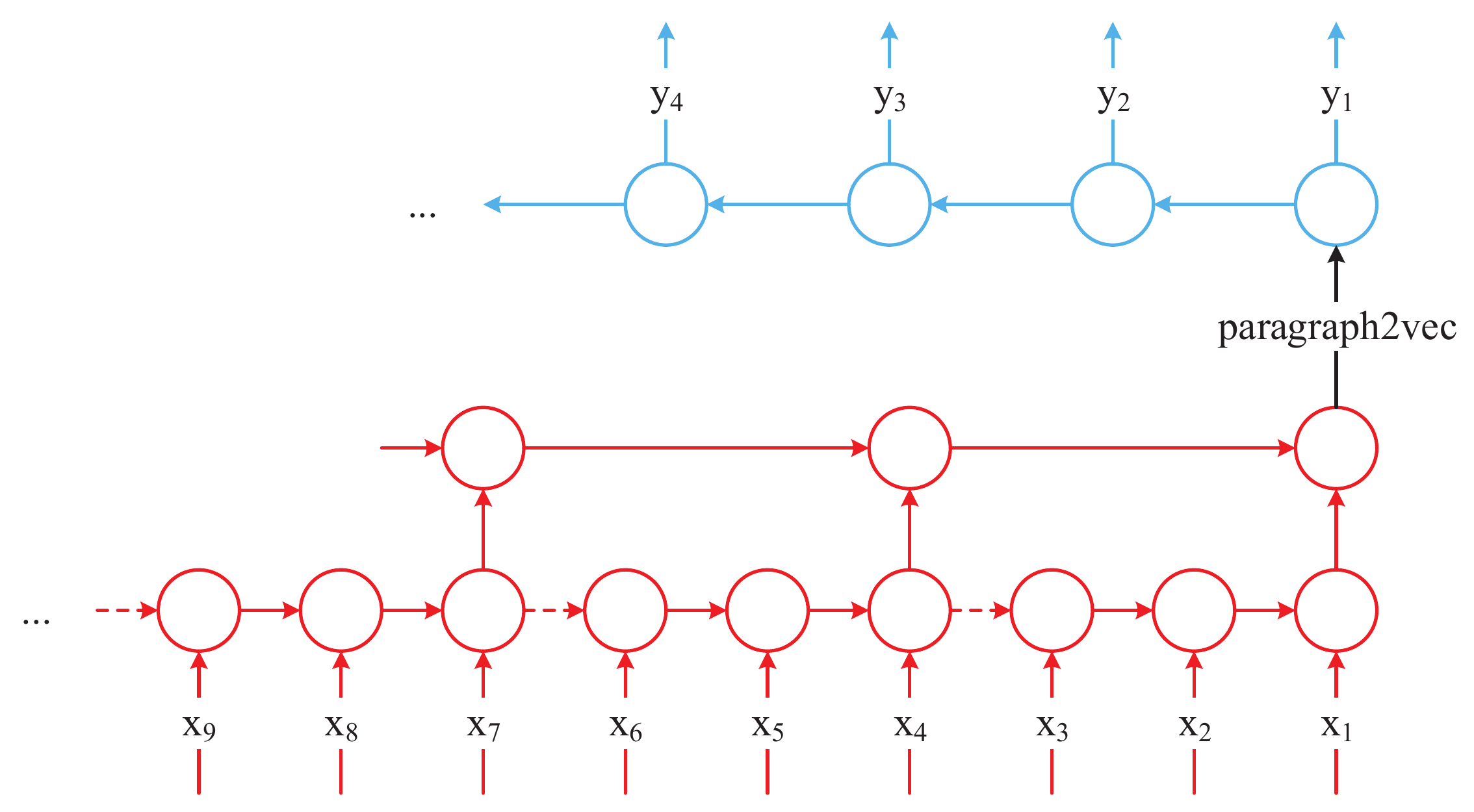}}
  \vspace{-.1in}
  \caption{The paragraph summarizer. The red and blue cells represent encoder and decoder respectively. The encoder inputs $x_i$ are the vector representation generated using sent2vec of the sentences in the paragraph. The decoder outputs $y_i$ are the words in summarized text. The intermediate vector in black (paragraph2vec) is vector encoded paragraph. 
  }
\label{fig:summarize}
\end{figure}

Each sentence in a paragraph can be represented by a vector using the method described in Section \ref{subsec:paraphrase}. These vectors $x_i, i \in \{1...T_x\}$ are fed into a hierarchical encoder \cite{pan2016hierarchical}, and then the summarized text is generated word by word in an RNN decoder, as shown in Fig. \ref{fig:summarize}. We first divide all $T_x$ vectors into several chunks $(x_1, x_2,... ,x_n)$, $(x_{1+s}, x_{2+s},... ,x_{n+s})$, ..., $(x_{T-n+1}, x_{T-n+2},...,x_T)$, where $s$ is the stride and it denotes the number of temporal units adjacent chunks are apart. For each chunk, a feature vector is extracted using a LSTM layer and fed into the second layer. Each feature vector gives a proper abstract of its corresponding chunk. We also use LSTM units in the second layer to build the hierarchical encoder. The first LSTM layer serves as a filter and it is used to explore local temporal structure within subsequences. The second LSTM learns the temporal dependencies among subsequences. As a result, the feature vector generated from the second layer, which is called ``paragraph2vec'', summarizes all input vectors extracted from the entire paragraph. Finally, a RNN decoder converts ``paragraph2vec'' into word sequence $(y_1, y_2, ...)$, forming a summarized sentence. 

We integrate a soft attention mechanism \cite{bahdanau2014neural} in the hierarchical encoder. 
The attention mechanism allows the LSTM to pay attention to different temporal locations of the input sequence.  When the input sequence and the output sequence are not strictly aligned, attention can be especially helpful.

\section{Experimental Results}
\label{sec:result}

\subsection{Datasets}

\textbf{Visual Caption Datasets}
There are numerous datasets with multiple captions for images or videos. For example, MSR-VTT dataset \cite{xu2016msr} is comprised of 10,000 videos with 20 sentences each describing the videos. The 20 sentences are paraphrases since all the sentences are describing the same visual input. We form pairs of these sentences to create input-target samples. Likewise, MSVD \cite{chen2011collecting}, MS-COCO \cite{Lin2014}, and Flickr-30k \cite{young2014image} are used. Table \ref{dataset} lists the statistics of datasets used. 5\% of captions from all datasets was held out as a test set. In total, we created over 10M training samples.

\vspace{-.2in}
\begin{table}[!ht]
\centering
\caption{Sentence pairs statistics in captioning datasets.}
\label{dataset}
\begin{tabular}{c|cccc}
\hline\hline
             & MSVD     & MSRVTT & MSCOCO    & Flickr  \\ \hline
\#sent       & 80K   & 200K & 123K    & 158K \\
\#sent/samp. & $\sim$42 & 20      & 5          & 5       \\
\# sent pairs& 3.2 M    & 3.8 M   & 2.4 M      & 600 K   \\
\hline \hline
\end{tabular}
\end{table}

\noindent
\textbf{The SICK dataset}
We use the Sentences Involving Compositional Knowledge (SICK) dataset \cite{marelli2014sick} as a test set for sentence paraphrasing task. It consists about 10,000 English sentence pairs, which are annotated for relatedness by means of crowd sourcing techniques. The sentence relatedness score (on a 5-point rating scale) for each sentence pair is provided and meant to quantify the degree of semantic relatedness between sentences.

\noindent
\textbf{TACoS Multi-Level Corpus}
We extract training pairs for the paragraph summarization task from TACoS Multi-Level Corpus \cite{rohrbach14gcpr}. This dataset provides coherent multi-sentence descriptions of complex videos featuring cooking activities with three levels of detail: ``detailed'', ``short'' and ``single sentence'' description. There are 143 training and 42 test video sequences with $\sim$20 annotations for each of the description levels in each sequence.

\subsection{Training Details}

\noindent
\textbf{Sentence Paraphrasing}
We trained the model as described in Section \ref{subsec:paraphrase} on the Visual Caption Datasets. The word embedding is initialized using GloVe \cite{pennington2014glove}. The number of units per layer in both encoder and decoder are empirically set to $300$ and $1024$. We generate two sets of vocabularies with size $20k$ and $50k$. Stochastic gradient descent is employed for optimization, where the initial learning rate and decay factor are set to $0.0005$ and $0.99$, respectively. 

\noindent
\textbf{Paragraph Summarization}
In this task we summarize ``detailed'' description to ``single sentence'' in TACoS Multi-Level Corpus. We select detailed descriptions with less than 20 sentences. There are total of 3,176 samples, 2,467 are used for training and 709 are used for testing. We employed the hierarchical architecture described in Section \ref{subsec:summarize} with stride $s$ of 5. 20 feature vectors (short paragraphs are zero padded) are fed into the model, with each vector's sentence representation extracted from our paraphrasing model. To make the model more robust, soft attention is used between each layer. During training, we use learning rate $0.0001$ and Adam optimizer. All the LSTM cells are set to 1024 units, except the one in sentence generation layer which is set to 256 units.

\begin{figure*}[!ht]
  \centering
  \begin{minipage}[b]{.3\linewidth}
    \centering
    \frame{\includegraphics[width=.9\textwidth]{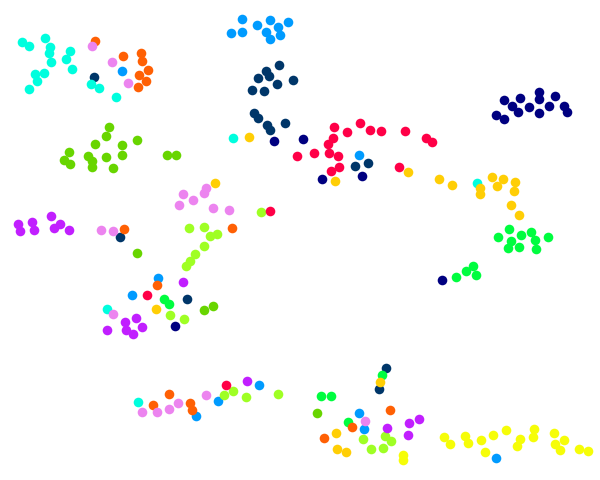}}
    \centerline{(a)}
  \end{minipage}
  \begin{minipage}[b]{.3\linewidth}
    \centering
    \frame{\includegraphics[width=.9\textwidth]{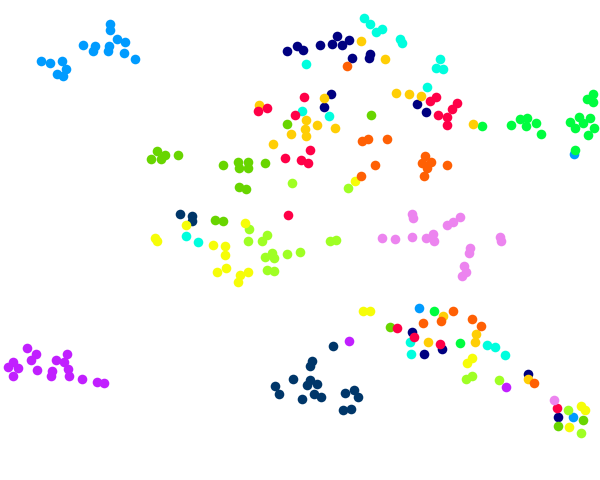}}
    \centerline{(b)}
  \end{minipage}
  \begin{minipage}[b]{.3\linewidth}
    \centering
    \frame{\includegraphics[width=.9\textwidth]{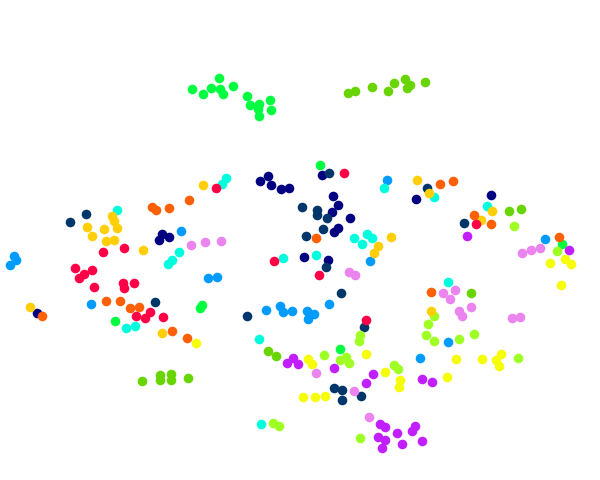}}
    \centerline{(c)}
  \end{minipage}
  \vspace{-.1in}
\caption{t-SNE visualizations of single sentence descriptions of a subset of all sequences on TACoS Multi-Level Corpus. (a) Our sent2vec; (b) Skip-thoughts and (c) Skip-gram. Points are colored based on their sequence IDs. There are $\sim$20 different annotations for each sequence. (Best viewed in color.)}
\label{summarize_tsne}
\end{figure*}

\subsection{Sentence Paraphrasing}
\label{subsec:experiment_paraphrase}

Given a reference sentence, the objective is to produce a semantically related sentence. 
The paraphrasing model was trained on Visual Caption Datasets and evaluated on the SICK dataset, without any fine-tuning. 
The results are shown in Table \ref{semantic_testresult}. The evaluation metrics for this experiment are Pearson's $r$, Spearman's $\rho$ and Mean Squared Error(MSE). We use the same setup used by \cite{socher} for calculation of these metrics.


\vspace{-.2in}
\begin{table}[!ht]
\centering
\caption{Test set results on the SICK semantic relatedness task, where 300, 1024 denote the number of hidden units and 20k, 50k denote the size of the vocabulary. $r$ and $\rho$ are Pearson's and Spearman's metric respectively.}
\label{semantic_testresult}
\begin{tabular}{c|ccc}
\hline\hline
        & $r$    & $\rho$    & MSE  \\ \hline
sent2vec(300,20k)        &0.7238    &0.5707    &0.4862   \\
sent2vec(300,50k)       &0.7472    &0.5892    &0.4520   \\ 
sent2vec(1024,50k)        &0.6673    &0.5285    &0.5679   \\
\hline\hline
\end{tabular}
\end{table}



\begin{figure}[!ht]
\begin{minipage}[b]{0.48\linewidth}
    \includegraphics[width=\linewidth, height=0.95\linewidth]{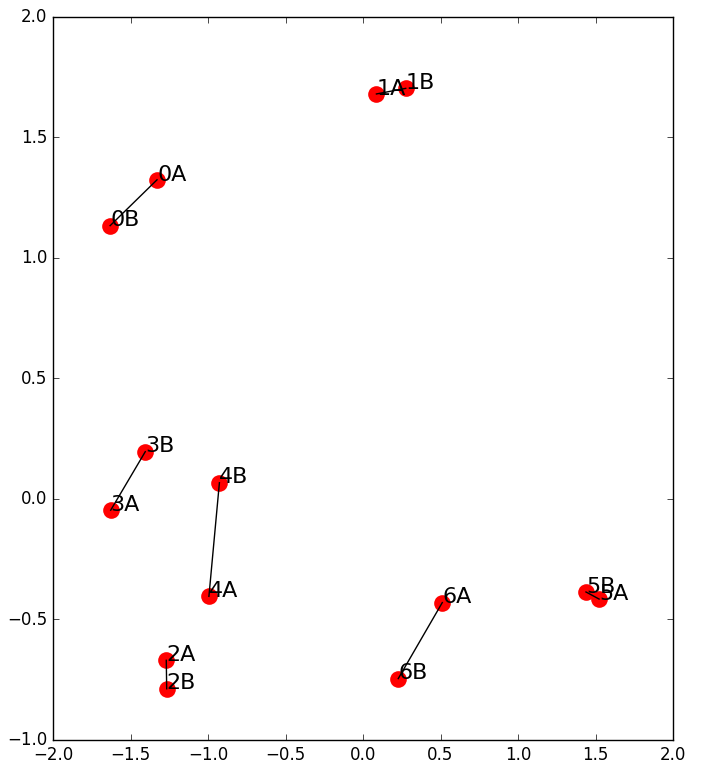}
\end{minipage}
\begin{minipage}[b]{0.48\linewidth}
    \tiny
    0A: the young boys are playing outdoors and the man is smiling nearby\\
    0B: a group of kids is playing in a yard and an old man is standing in the background\\
    1A: a brown dog is attacking another animal in front of the man in pants\\
    1B: a brown dog is helping another animal in front of the man in pants\\
    2A: two people are kickboxing and spectators are watching\\
    2B: two people are fighting and spectators are watching\\
    3A: kids in red shirts are playing in the leaves\\
    3B: three kids are jumping in the leaves\\
    4A: a little girl is looking at a woman in costume\\
    4B: the little girl is looking at a man in costume\\
    5A: a woman is removing the peel of a potato\\
    5B: a woman is peeling a potato\\
    6A: five children are standing in front of a wooden hut\\
    6B: five kids are standing close together and one kid has a gun
\end{minipage}
\caption{Some paraphrase sentence pairs are represented by our sent2vec and then projected into 2D space using PCA. Each point represents a sentence in SICK dataset and the corresponding sentence is shown on the right.}
\label{fig:PCA}
\end{figure}

In order to visualize the performance of our method, we applied PCA to the vector representation. Fig. \ref{fig:PCA} visualizes some of the paraphrase sentence pairs in the SICK dataset. Representations are sensitive to the semantic information of the sentences since pairwise sentences are close to each other. For example, point 2A and 4A are close because ``watching'' and ``looking'' are semantically related.


The semantic relatedness and grammar correctness are verified by human generated scores. Each score is the average of 32 different human annotators. Scores take values between 1 and 5. A score of 1 indicates that the sentence pair is not at all related or totally incorrect syntax, while a score of 5 indicates they are highly related or grammatically correct. The sentences in human evaluation come from the Visual Caption and SICK test sets. The human evaluated scores of most sentence pairs are inversely proportional to the Euclidean distance between the vector representation of the corresponding sentences.

\subsection{Text Summarization}

We now show that in addition to paraphrasing,  sent2vec is useful for text summarization. We use the TACoS Multi-Level Corpus \cite{rohrbach14gcpr}. Sentences from detailed descriptions of each video sequence are first converted into vectors using our model. These vectors are fed into our summarizer described in Section \ref{subsec:summarize}. The performance of the summarized text is evaluated based on the metric scores and compared to skip-thoughts and skip-gram. Note that skip-gram is used as the frequency-based average of word2vec for each word in the sentence. As shown in Table \ref{summarize_result}, the scores generated by our model are very close and comparable to the benchmark skip-thoughts. This result is reasonable since the dataset used in training sent2vec are all from captions. The styles and topics of the sentences in this dataset are limited. However, the approach of forming sentence paraphrasing pairs and representing sentences using vectors are valid.

\vspace{-.1in}
\begin{table}[!ht]
\centering
\caption{Evaluation of short to single sentence summarization on TACoS Multi-Level Corpus using vectors from sent2vec, skip-thoughts, and skip-gram respectively.}
\label{summarize_result}
\begin{tabular}{c|ccc}
\hline\hline
        & sent2vec    &  skip-gram   &  skip-thoughts \\ \hline
BLEU-1        & 0.479   & 0.514   & 0.520  \\ 
BLEU-2        & 0.342   & 0.378   & 0.392  \\
BLEU-3        &  0.213  & 0.245   & 0.276  \\
BLEU-4        &  0.144  & 0.173   & 0.206  \\
METEOR        &  0.237  & 0.250   & 0.253  \\
ROUGE-L       &  0.48  & 0.509   & 0.522  \\
CIDEr         &   1.129 & 1.430   & 1.562  \\ \hline\hline
\end{tabular}
\end{table}



Fig. \ref{summarize_tsne} shows t-SNE \cite{maaten2008visualizing} vector representation using (a) our sent2vec, (b) skip-thoughts and (c) skip-gram of randomly selected $\sim$15 test sequences. In these plots, each point represents a single sentence. Points describing the same video sequence should be clustered. Points with the same color are nicely grouped in sent2vec visualization.

\section{Conclusion}
\label{sec:conclusion}
We showed the use of a deep LSTM based model in a sequence learning problem to encode sentences with common semantic information to similar vector representations. The presented latent representation of sentences has been shown useful for sentence paraphrasing and document summarization. We believe that reversing the encoder sentences helped the model learn long dependencies over long sentences. One of the advantages of our simple and straightforward representation is the applicability into a variety of tasks. Further research in this area can lead into higher quality vector representations that can be used for more challenging sequence learning tasks.

\bibliographystyle{IEEEbib}
\bibliography{refs}

\end{document}